# OmniTFT: Omni Target Forecasting for Vital Signs and Laboratory Result Trajectories in Multi Center ICU Data

Wanzhe Xu[1], Yutong Dai[1], Yitao Yang[1], Martin Loza[2], Weihang Zhang[1], Yang Cui[1], Xin Zeng[1], Sung Joon Park[2,3], and Kenta Nakai[1,2*]

[1]Department of Computational Biology and Medical Science, The University of Tokyo, Kashiwa, Japan,

[2]Human Genome Center, The Institute of Medical Science, The University of Tokyo, Tokyo, Japan

[3]Department of Frontier Research and Development, Kazusa DNA Research Institute, Chiba, Japan

*Kenta Nakai (knakai@ims.u tokyo.ac.jp)

## Abstract

Accurate multivariate time-series prediction of vital signs and laboratory results is crucial for early intervention and precision medicine in intensive care units (ICUs). However, vital signs are often noisy and exhibit rapid fluctuations, while laboratory tests suffer from missing values, measurement lags, and device-specific bias, making integrative forecasting highly challenging. To address these issues, we propose OmniTFT, a deep learning framework that jointly learns and forecasts high-frequency vital signs and sparsely sampled laboratory results based on the Temporal Fusion Transformer (TFT). Specifically, OmniTFT implements four novel strategies to enhance performance: sliding window equalized sampling to balance physiological states, frequency-aware embedding shrinkage to stabilize rare-class representations, hierarchical variable selection to guide model attention toward informative feature clusters, and influence-aligned attention calibration to enhance robustness during abrupt physiological changes. By reducing the reliance on target-specific architectures and extensive feature engineering, OmniTFT enables unified modeling of multiple heterogeneous clinical targets while preserving cross-institutional generalizability. Across forecasting tasks, OmniTFT achieves substantial performance improvement for both vital signs and laboratory results on the MIMIC-III, MIMIC-IV, and eICU datasets. Its attention patterns are interpretable and consistent with known pathophysiology, underscoring its potential utility for quantitative decision support in clinical care.

# Introduction

Electronic health record (EHR) data are critical for diagnostic decision-making and precision medicine[1]. These data originate from distinct sources and exhibit diverse temporal and statistical patterns, reflecting the complexity of clinical practice. In the intensive care unit (ICU), patient conditions recorded in EHRs can change rapidly over short time scales ranging from minutes to hours, posing a challenge for clinicians to accurately predict future physiological states based on experience alone[2,3].

In recent years, an increasing number of researchers have employed machine learning algorithms to forecast trends in EHRs[4]. Their efforts primarily focus on two aspects: (1) Prediction of vital sign trajectories, such as LSTM[5], GRU[6], or traditional regression models[7], which are simple to implement but susceptible to irregular sampling and missing mechanisms in cross-center scenarios[8,9]. Among recent models, Prophet[10] uses seasonal trends and residual decomposition, while TSMixer[11] employs an MLP mixture of all-time combined features to better handle irregular sampling and covariate utilization. However, both approaches struggle to capture asynchronous dependencies and learn complex relationships among covariates, which limits their ability to detect the abrupt changes often observed in ICU patients. (2) Prediction of laboratory results. For instance, Nephrocast[12] can effectively learn task-specific relationships and demonstrate good performance in creatinine forecasting. Its extended version, Nephrocast-V[13], further predicts vancomycin trough concentrations 48 hours in advance using a deep model combining LSTM layers and multi-head attention. However, both of them focus on a single laboratory indicator with a short lead time and have limited support for joint learning of high-frequency vital signs or robust calibration during cross-hospital generalization.

To address both vital signs and laboratory results within a single framework, recent studies have explored the Temporal Fusion Transformer (TFT) for heterogeneous, multivariate forecasting. Vanilla TFT[14,15] implements a combination of local recursive encoding and self-attention to handle complex relationships between multiple covariates; however, it has limited capacity to reduce sensitivity to high-frequency variability in clinical time series. This may create a temporal lag in the results, thereby deferring subsequent clinical alert workflows.

In addition to the drawbacks mentioned above, current feature selection methods still rely on clinician expertise. This manual process can overlook rare variables that may affect predictions, leading to inaccurate identification of emergency risk[16,17]. To overcome these limitations, we propose OmniTFT, a deep-learning framework that jointly predicts vital signs and laboratory results. OmniTFS is trained on all bedside monitoring data and laboratory measurements, eliminating the need for manual feature selection. Compared to several existing methods, OmniTFT achieved more than a 20% improvement across multiple prediction tasks, demonstrating superior predictive capability. Through rigorous validation using cross-validation, 10-random seed training, and external cohort validation, OmniTFT consistently outperformed baseline models across various patient populations and clinical settings. The significant predictive accuracy indicates the potential of OmniTFT to enable early detection of clinical deterioration and to support proactive interventions in critical care.

# Result

**Overview of OmniTFT**

To prepare inputs for OmniTFT, we selected a dataset for training and internal validation, along with two external validation cohorts (Figure 1A). Then, to capture plateaus and changepoints, OmniTFT applies a two-state Hidden Markov Model (HMM, see Methods) to the differenced signal, yielding stable and volatile labels[18]. These labels then guide a sliding-window sampler that forms feature-combination tokens under a balanced scheme for plateau and transition periods (Figure 1B).

Next, to enhance the sensitivity of our model to changepoints in clinical time series, frequency-aware regularization is applied to mitigate class imbalance between stable and volatile periods[19]. After stabilization, past measurements are encoded using two LSTM layers, and static covariates are used for decoding future steps. Gated residual networks with hierarchical variable selection and cross-masked attention capture dependencies across different temporal scales, while bump-aware alignment anchors focus on true physiological changes rather than noise (Figure 1C).

OmniTFT is optimized with a hierarchical, regularized objective to enhance robustness and interpretability[20]. The trained model produces feature-importance scores from variable selection, temporal-attention heatmaps, and per-sample attributions.

**Validation of model performance**

To evaluate predictive accuracy and cross-cohort generalization, we benchmarked five vital signs (blood pressure, heart rate, SpO2, Respiratory rate, and Temperature) and two laboratory results (Creatinine and Lactate) across three ICU datasets: MIMIC-IV[21] as a test set, MIMIC-III[22] and eICU[23] for validation. We also included a derived oxygenation index, the SF ratio, calculated as SpO2 divided by $FiO_2$ using bedside pulse oximetry and the recorded fraction of inspired oxygen. We chose to model the SF ratio because it non-invasively tracks hypoxemia severity as a validated surrogate for the $PaO_2/FiO_2$ ratio; SF also shows prognostic value for mortality in ARDS[24]. To prevent overlap between patients in MIMIC-III and MIMIC-IV, we selected MIMIC-IV patient data from 2008 to 2019 and MIMIC-III patient data from 2001 to 2005. Table A-1 summarizes the error performance of the chosen vital signs and laboratory results across the datasets. Each cell provides paired values for mean absolute error (MAE) and mean absolute percentage error (MAPE)[25]. Supplementary Information Tables S2-1 to S2-16 provide sub label and sub seed results for additional auxiliary metrics: RMSE, RMBE, P10 Coverage, P10 Pinball, P90 Coverage, and P90 Pinball[26,27].

The magnitude and ranking of prediction errors were largely consistent across the three datasets, with no significant differences observed in external validation. Relative errors for vital signs generally remained within the single-digit percentage range, whereas absolute errors for body temperature were minimal, remaining within sub-degree ranges. Creatinine and lactate exhibited higher percentage errors due to low values and intermittent sampling. The SF ratio showed larger absolute errors but lower relative errors. Overall, these results indicate that the model is robust to variations in sampling frequency and population shifts.

**Comparison of OmniTFT with Other Methods**

We compared our methodology with baselines (Vanilla TFT, LSTM, Prophet, TSMixer, GRU, and Nephrocast; Table A-2 and Figure 2). OmniTFT achieved the best absolute and relative error performance in six of eight metrics, covering both vital signs and laboratory parameters with a single model. Compared with statistical baselines, OmniTFT demonstrated significant reduction in absolute error for labels such as blood oxygen saturation and body temperature. This improvement likely stems from its ability to capture nonlinear temporal dependencies that traditional methods cannot represent.

Building on these foundational improvements, OmniTFT also outperformed the Vanilla TFT architecture. For example, the mean absolute error for heart rate and blood pressure decreased by nearly 70%, while creatinine predictions improved by nearly 90%, demonstrating the effectiveness of our proposed regularization layers in handling complex multivariate inputs and clinical noise. Across the three validation cohorts, OmniTFT maintained similar absolute error levels and, for most targets, performed favorably compared to other baselines. Taken together, these results indicate that OmniTFT provides more accurate predictions across a wide range of vital-sign and laboratory tasks, supporting potential clinical application.

**Representative Forecast Trajectories**

We examined individual forecast trajectories to assess the model's behavior in clinical scenarios. Figure 3 shows forecasts for eight indicators, each of subfigure presents a patient whose MAE for that specific indicator is closest to the cohort mean, thereby illustrating a typical forecasting case[28]. Overall, median forecasts closely matched the true future trends, and uncertainty bands expanded smoothly with the forecast horizon, indicating that the intervals are reliably calibrated and remain stable within a two-hour window. For slowly changing laboratory indicators such as creatinine and temperature, the intervals were narrow and track monotonically. For high-frequency vital signs (heart rate, respiratory rate, $SpO_2$, and blood pressure), the model responded rapidly to sudden changes near t = 0 and gradually regressed after brief fluctuations. Taken together, these results demonstrate that OmniTFT effectively captures dynamic trends and accurately predicts both vital-sign and laboratory indicators.

**Model Attention in OmniTFT Across Different Time States and Feature Combinations**

To get insights into the internal decision-making process of OmniTFT, we conducted a comprehensive analysis of the self-attention mechanism. Figure 4A shows the attention scores for the 20 features with the highest weights across eight indicators. Certain features, such as 24-h creatinine, appear as zero in the creatinine prediction because they were masked during preprocessing to prevent label leakage.

We further analyze the top attention scores described above by ranking them according to their relative contribution to indicator predictions (Figure 4B). Blood pressure and heart rate showed the highest concentration of top-ranked features, followed by body temperature and lactate, while other indicators showed more dispersed attention. These differences reflect indicator-specific feature effects, where for certain indicators, such as blood pressure and heart rate, the model focuses more on specific key markers, whereas for others, such as creatinine, attention is distributed across multiple features. This differentiated and cross-domain allocation reveals a multi-system integration strategy.

When examining the physiological attribution of all high-attention features, we found that they were distributed across eight organ systems rather than clustered in a single domain (Figure 4C). OmniTFT captures inter-organ interactions and comorbidity patterns by integrating multi-system features, rather than relying solely on direct indicators within a single system. To verify the robustness of these attention patterns across tasks, we analyzed the variation of the attention values across signs (Figure 4D). In cross-validation, 72.1% of the features had a coefficient of variation below 0.5 (median CV: 0.307), indicating that the complex patterns identified in Figure 4C are unlikely due to chance.

Combining the findings from the analyses above, OmniTFT adopts a hierarchical decision trajectory. First, it scans clinically relevant features across physiological systems to identify organ interactions and comorbidity patterns. Next, it selectively strengthens the weights of specific indicators according to the prediction task, and finally forming a task-specific attention distribution. It is important to note that the attention weights indicate observation priority, not causal importance[29,30].

**Feature Importance of OmniTFT Under Different Sampling Methods on Global Variables—SHAP Analysis**

To quantify individual feature contributions beyond the general model attention presented in the previous section, we performed SHAP analysis[31]. A feature-target heatmap was created to quantify the marginal contribution of each feature to the final prediction, revealing distinct patterns (Figure 5A): CK consistently displayed high SHAP values across multiple targets, whereas the alveolar-arterial gradient showed high importance for lactate, consistent with previous studies[32,33]. To further explore feature importance at the sample level, we performed single-target sample SHAP analysis. Figure 5B illustrates an example of heart rate results, where the SHAP distribution of individual patients showed a concentration of certain features, such as CK, in the negative SHAP region. However, the associations varied across patients, with some exhibiting positive SHAP values.

To investigate the scope of feature contributions, we compared the mean SHAP values across target coverage (Figure 5C). Plotting target coverage against average SHAP revealed that CK, ferritin, and LD appeared in ≥6 targets, while features such as spontaneous respiration and protein had narrower coverage. To further assess the stability of these patterns, we quantified the variability of SHAP values across samples (Figure 5D), with CK and ferritin displaying wide distributions, indicating their patient-specific importance. Finally, we analyzed the directionality of average SHAP values across patients (Figure 5E), with CK and ferritin displaying relatively symmetrical bidirectional bars, supporting our previous observation of patient-specific importance. In summary, the SHAP analysis demonstrates that OmniTFT does not rely on a fixed set of universal predictors but assigns feature importance based on the specific condition of the patient and the prediction task.

## Discussion

Multi-center ICU time series often switch between stable and unstable states. Under these conditions, the attention weights and variable selection scores of the vanilla TFT model can be unstable, and simple feature selection can lead to omitted variable bias[34,35]. Additionally, long-tailed categorical covariates may cause overfitting of rare categories in the embedding, increasing variance and exacerbating domain shift[36]. These issues motivate the development of regularization and alignment extensions that prioritize both stability and accurate recognition.

In this study, we propose OmniTFT, a unified deep learning framework for the joint modeling of vital signs and laboratory results. OmniTFT integrates balanced sampling, frequency-aware embedding shrinkage, hierarchical variable selection, and influence-aligned attention calibration into a cohesive architecture. This framework offers several key advantages that support its potential for clinical application. First, OmniTFT demonstrates consistent performance across diverse patient cohorts, indicating strong generalizability. Second, the attention mechanisms provide interpretable insights into the feature relationships underlying prediction, enabling transparent and clinically meaningful model outputs. Finally, the modular design of OmniTFT allows incremental extension as new biomarkers or data sources become available. This adaptability makes OmniTFT particularly suitable for resource-constrained clinical environments where full model retraining is impractical.

Our interpretability analysis revealed patterns that align well with established clinical knowledge. For example, feature importance analysis highlighted that cardiac markers dominate blood pressure prediction, while renal and inflammatory markers were prioritized for creatinine prediction. Notably, NT-proBNP received the highest attention in blood pressure prediction, consistent with the use of natriuretic peptides for hemodynamic risk stratification[37]. Similarly, the CK-MB index emerged as a key contributor to SF ratio predictions. In addition to task-specific variables, we identified several global biomarkers, including creatine kinase, ferritin, and lactate dehydrogenase, exert broad influence across multiple prediction targets. These features demonstrated stable effects across diverse tasks. In contrast, features such as spontaneous respiration and protein exhibit greater inter-patient variability, with their impact differing across individuals. This highlights that OmniTFT adaptively leverages heterogeneous features to generate precise, patient-specific predictions. The consistency of these patterns on the validation set supports the conclusion that OmniTFT captures physiologically meaningful dependencies rather than dataset-specific correlations.

Despite the advantages described above, several challenges remain. First, predictions for sparsely sampled laboratory test results remain suboptimal compared to specific models. We suspect that sampling leaves limited joint samples between features and the target within the window, weakening the available signal. A larger patient cohort is likely to mitigate this limitation[38]. Second, OmniTFT requires a large feature set to learn within-window feature combinations across diverse physiological contexts, where a reduced input feature availability would negatively affect prediction accuracy. This effect is likely due to limiting the model's ability to capture conditional dependencies[39].

We envision that future developments could explore multimodal integration to capture physiological dynamics beyond structured electronic health record data. Incorporating radiology images, echocardiographic videos, and continuous waveforms may provide complementary information across multiple dimensions, while cross-modal alignment mechanisms would be required to reconcile differing semantic spaces and sampling frequencies. In addition, meta-learning and transfer learning approaches could reduce the sample requirements for rare diseases or resource-constrained institutions, lowering barriers to adoption while maintaining broad applicability. Together, these advances have the potential to further enhance predictive models; however, careful validation using data from multiple modalities is essential to ensure robustness across diverse clinical settings and patient populations.

Importantly, OmniTFT demonstrated both transferability and interpretability, suggesting that unified temporal modeling of heterogeneous ICU data represents a promising avenue for quantitative decision support in critical care.

## Methods

### Data Description

This study utilized three independent, publicly available critical care databases for model development and external validation. The Medical Information Mart for Intensive Care IV (MIMIC IV) v2.2 was employed as the primary training dataset, a large, freely accessible electronic health record database sourced from Beth Israel Deaconess Medical Center that contains comprehensive clinical data from patients admitted between 2008 and 2019.

For external validation, two independent cohorts were utilized. First, we utilized the MIMIC III v1.4 database, a large, single-center database comprising information relating to over 40,000 patients admitted to critical care units at the same institution between 2001 and 2012. To ensure temporal independence and avoid data overlapping with the training cohort, we exclusively utilized patient records from 2001 to 2005. Second, the eICU Collaborative Research Database (eICU-CRD) v2.0 is a multicenter intensive care unit database comprising health data from over 200,000 admissions across 208 hospitals in the United States between 2014 and 2015. All three databases use de-identified patient data. As this study utilized only publicly available data, no additional Institutional Review Board (IRB) approval was required.

### Data preprocessing

We retained patients from MIMIC IV with one of three common ICU conditions: Sepsis, ARDS, or AKI[40–43]. We also retained patients from MIMIC III and eICU who had one of these three conditions. After data preprocessing, MIMIC IV included 13,277 patients, MIMIC III included 4,188 patients, and eICU included 23,238 patients. We used a filter based on the original MIMIC IV item dictionary to remove nonphysiological records while preserving core vital signs. Manual administrative tags were discarded for laboratory events. Our final data input consisted of de-identified patient IDs and examination timestamps, a numeric timestamp based on the number of hours from the patient's current measurement date to the hospital admission date, the patient's age, height, weight, race, hospital type, 172 laboratory results, and bedside notes. All input data types, units, and ranges are listed in Supplementary Information Tables S-1 through S-3. We removed data from patients with more than 80% missing values for static or observed features. For the remaining missing values, median or forward imputation was performed based on the feature attributes[44], with a maximum allowable gap of 6 hours[45]. If the first observation was missing, it was deleted[46].

We divided the MIMIC IV training, validation, and test sets into a 7:2:1 ratio based on patient ID. We used 10-minute resampling for vital signs and the SF ratio[47], 2-hour resampling for laboratory results to reflect sparse data collection[48]. Vital signs used 12-hour historical data to predict the next 2 hours, while laboratory results used 24-hour data to predict the next 24 hours[6,49]. To balance the training data, we adopted dynamic window sampling that pairs time segments with sharp vital sign fluctuations against segments with relatively stable values. This pairing ensures equal representation of both dynamic and stable physiological states, preventing the model from being biased toward one pattern. The names, units, and statistical information for all input data are described in Supplementary Tables S-1 and S-2.

**Vanilla TFT**

We describe Vanilla Temporal Fusion Transformer (TFT) as a comparison. The original Temporal Fusion Transformer uses a sequence encoder to model local temporal patterns and a self-attention mechanism to capture long-range dependencies. It encodes both static and time-varying covariates and employs a variable selection network and a gating mechanism to regulate information flow. It supports multi-temporal predictions and produces probabilistic outputs via a quantile loss.

**OmniTFT**

We propose OmniTFT, which extends the Vanilla TFT architecture based on the characteristics of ICU data. At the data level, OmniTFT first deploys a state-balanced sliding window sampler that dynamically adjusts the exposure to stable and volatile periods before the data enters the embedding layer. After sampling, the input flows through three regularities that act sequentially along the depth of the network to stabilize the predictions under different states. To address the limitations of Vanilla TFT, OmniTFT introduces three mechanisms. (1) In the embedding layer, frequency-aware embedding shrinkage prevents overfitting to rare categorical features by shrinking low-frequency embeddings to the learned prior. (2) In the temporal encoder, hierarchical group selection performs decisive variable selection by minimizing the entropy within the learned variable groups. (3) In the decoder, shock-aligned attention calibration recalibrates the self-attention weights to focus on true state transitions rather than noisy fluctuations.

**State-balanced sliding window sampler**

First, align each patient's slices to a fixed time grid to generate a chronologically ordered target sequence $y_{1:n}$. We then slide windows of length $T = E + H$ with encoder length E and horizon length H over the sequence. For a window starting at index $t \in \{1, \ldots, n - T + 1\}$ the encoder segment is $y_{t:t+E-1}$ and the future segment is $y_{t+E:t+T-1}$. We determine the relationship between fluctuations and physiological change thresholds by calculating the future segment fluctuation score $S_t$ as:

$$S_t = \max_{\tau \in [t+E, t+T-1]} y_\tau - \min_{\tau \in [t+E, t+T-1]} y_\tau \tag{1}$$

The score retains the same unit as the target and is simply the max minus min over the future window. We then set a clinically reasonable cutoff $\delta$ for that target. If $S_t > \delta$, the segment is labeled as "volatile", otherwise the segment is labeled as stable.

**Hierarchical Constraint Training**

The training process can be marked as

$$\mathcal{L}_{\text{total}}(\theta) = \mathcal{L}_{\text{quantile}}(\theta) + \sum_{j \in \text{embed,group,shock}} \lambda_j \, C_j(\theta) \tag{2}$$

where $\theta$ means all trainable parameters. The first constraint acts on categorical embeddings at the input stage, the second acts on temporal variable selection weights after the embedding has been formed, and the third acts on self-attention after temporal fusion has produced the decoder-side representation. The penalty strengths reflect hierarchical depth $\lambda_{\text{shock}} = 10^{-1}, \quad \lambda_{\text{group}} = 10^{-2}, \quad \lambda_{\text{embed}} = 10^{-3}$

**Frequency Aware Embedding Shrinkage**

At the embedding entrance, the risk is that long-tail categories inject high-variance directions, since each category is granted a free embedding row even when evidence for that category is scarce. To counter this, we regulate the energy of each embedding row by the precision of its empirical frequency inside the mini batch[24]. Consider a categorical feature $i$ with vocabulary size $\mathcal{V}_i$ and embedding matrix $W_i \in R^{V_i \times H}$. Let $c_{i,k}$ be the batch count of the category $k$ and $p_{i,k} = c_{i,k} / \sum_{u=1}^{V_i} c_{i,u}$ is batch frequency. For category $k$ of feature $i$ with batch frequency $p_{i,k}$, define

$$R_{\text{embed}}^{(i,k)} = \frac{\| W_i[k,:] \|_2^2}{\sqrt{p_{i,k} + \varepsilon}}, \quad p_{i,k} = \frac{c_{i,k}}{\sum_{u=1}^{V_i} c_{i,u}} \tag{3}$$

We impose an $\ell_2$ penalty on each embedding row that is inversely weighted by the square root of its batch frequency. The factor $1/\sqrt{P_{i,k} + \varepsilon}$ reflects the classical $1/\sqrt{n}$ scaling of sampling uncertainty: rarer categories carry higher estimation variance and therefore receive stronger shrinkage[50]. The small constant $\varepsilon > 0$ prevents numerical blowup for missing categories and caps the maximum shrinkage for ultra-rare ones. Geometrically, the penalty is isotropic in the embedding space, while its strength is category adaptive through $p_{i,k}$, stabilizing long tail rows without unnecessarily constraining frequent categories.

$$C_{\text{embed}} = \frac{1}{I} \sum_{i=1}^{I} \left( \frac{1}{V_i} \sum_{k=1}^{V_i} R_{\text{embed}}^{(i,k)} \right) \tag{4}$$

Consistent with GradNorm[51], we first average $R_{\text{embed}}^{(i,k)}$ across the vocabulary of feature $i$ and then across features. The within feature average $1/\mathcal{V}_i \sum k$ prevents large vocabularies from dominating the loss simply because they contain more rows, the across feature average gives each categorical feature equal influence regardless of its cardinality. This normalization makes $\lambda_{\text{embed}}$ a single, interpretable global knob and yields a fair, invariant penalty: models are not penalized for having larger tables, only for hosting high variance rows. Here $I$ means the total number of categorical features

$$\nabla W_{i[k,:]} R_{\text{embed}}^{(i,k)} = \frac{2 W_i[k,\cdot]}{\sqrt{p_{i,k} + \varepsilon}} \tag{5}$$

The gradient acts exactly like category-specific weight decay, whose coefficient increases as $p_{i,k}$ decreases: updates remain aligned with the row vector, rare categories undergo larger decay steps that quickly damp noisy excursions, and frequent categories retain capacity due to small extra decay. Hence, frequent categories experience light shrinkage, while rare ones are pulled more strongly toward the origin, reducing variance at the source before any temporal processing.

**Hierarchical Variable Selection via Semantic Grouping**

After embeddings have been stabilized, the next risk is that many collinear channels diffuse temporal selection weight and confuse subsequent attention. We therefore compress selection at the level of semantic groups that are native to the Temporal Fusion Transformer formulation. Let $w_t^{hs} \in \mathbb{R}^{N_h}$ be the nonnegative weights assigned by the historical variable selection at time $t$ and let $w_t^{fut} \in \mathbb{R}^{N_f}$ be the nonnegative weights assigned to future known inputs. We introduce a fixed binary assignment matrix $G \in \{0,1\}^{3 \times N_h}$ with $G_{g,j} = 1$ if variable $j$ belongs to a group $g \in \{\text{unknown}, \text{known}, \text{observed}\}$. Aggregation and normalization are:

$$s_t^{hs} = G w_t^{hs} \in \mathbb{R}^3, \quad p_g^{hs}(t) = \frac{s_{t,g}^{hs}}{\sum_{g'=1}^{3} s_{t,g'}^{hs} + \varepsilon} \tag{6}$$

where $\varepsilon = 10^{-6}$ ensures numerical stability when group weights approach zero. The left relation aggregates channel scores into the three groups while preserving nonnegativity. The right relation normalizes the aggregated scores to the probability simplex, so that the subsequent constraint depends only on the distribution of mass across groups rather than on the absolute magnitudes of the scores. For the future side, only known exogenous variables are available by construction, while the unknown and observed groups are structurally absent during the forecast horizon. To preserve the same geometry and to allow a uniform constraint, we embed the future side into the same group space by letting

$$s_t^{fut} = \begin{bmatrix} 0 \\ 1^\top w_t^{fut} \\ 0 \end{bmatrix}, \quad p_t^{fut} = \frac{s_t^{fut}}{1^\top s_t^{fut}} \tag{7}$$

This construction enforces exact zeros for the structurally missing groups and places the entire future selection mass in the known entry, thereby maintaining comparable group distributions across the two sides and resulting in a well-posed entropy objective. With the two group distributions in hand, we quantify dispersion by Shannon entropy[52], and we minimize its time average across both sides

$$\mathcal{L}_{group} = \frac{1}{2}\left(\frac{1}{T}\sum_{t=1}^{T} H\left(p_t^{hs}\right) + \frac{1}{T}\sum_{t=1}^{T} H\left(p_t^{fut}\right)\right), \quad H(p) = -\sum_{g=1}^{3} p_g \log p_g \tag{8}$$

The simplex constraint in equations three and four makes entropy an appropriate potential since it attains its minimum at the vertices where a single group dominates. This will minimize the time-averaged entropy, thereby encouraging a focus on one informative group at each time and suppressing spurious oscillations in group attention. With input embeddings stabilized, we next address the issue of variable selection redundancy.

**Shock Aligned Calibration of Self Attention**

We ensure that decoder attention responds to genuine changes in the latent state rather than to noise. Let $A^{(m)}(t,\tau)$ be the causal self-attention weight from the target time $t$ to past time $\tau \leq t$ for head $m$. We reduce inter-head variance and improve identifiability by averaging the heads to obtain a single interpretable attention surface

$$\bar{A}(t,\tau) = \frac{1}{M}\sum_{m=1}^{M} A^{(m)}(t,\tau) \tag{9}$$

The goal is variance reduction and identifiability rather than mere simplification. Although all heads share the same expectation, head-specific fluctuations act as nuisance variation. Averaging cancels these idiosyncrasies and yields a statistically stable attention field. Such stability is essential before compressing a two-dimensional surface into a one-dimensional local retro mass, because uncontrolled compression can amplify upstream noise. We then compress this two-dimensional surface along causal diagonals into a one-dimensional local retro mass that captures the amount of recent context used by the decoder[53,54]. The compression is performed by summing off-diagonal weights within a short

causal window of width $W$

$$a_t = \sum_{k=1}^{W} \bar{A}(t, t-k) \tag{10}$$

We set $W = 3$ to capture immediate temporal dependencies relevant for acute physiological events while avoiding noise from distant correlations. Then aggregate the attention of temporally adjacent positions to obtain the short term lookback strength; to avoid trivial terms, self-loops of $t$ pointing to $t$ are excluded. A small window isolates reliance on the most recent history, matching the time scale of acute physiological shocks. This geometric projection converts a surface into a time series that can be aligned with a state variability index. The companion signal to which this local mass should respond is a measure of the instantaneous variability of the decoder side representation. Let $v_t \in \mathbb{R}^H$ denote that representation. The most direct index of abrupt change that requires no change point labels is the Euclidean first difference of the representation, defined as $s_t^{(1)} = \| v_t - v_{t-1} \|_2$ where $v_t$ means the decoder hidden state at time $t$. First order differencing is chosen over higher orders because acute episodes primarily manifest as magnitude jumps which are captured sensitively by $s_t^{(1)}$. This proxy avoids explicit change point labels, integrates cleanly with end to end training, and matches the temporal resolution of the local retro mass[55].

$$\tilde{s}_t^{(1)} = \frac{s_t^{(1)} - \mu_s}{\sigma_s}, \quad \tilde{a}_t = \frac{a_t - \mu_a}{\sigma_a} \tag{11}$$

Because $a_t$ and $s_t^{(1)}$ have different scales and units, naive alignment would be dominated by magnitude instead of temporal co-variation. We apply per-sequence standardization to both. This places both series on a comparable, unit free scale and ensures that the subsequent loss penalizes discrepancies in relative dynamics rather than absolute size. On the set of decoder indices $\mathcal{D}$ we then implement soft alignment by minimising the average squared gap

$$\mathcal{L}_{\text{shock}} = \frac{1}{|\mathcal{D}|} \sum_{t \in \mathcal{D}} \left( \tilde{a}_t - \hat{s}_t^{(1)} \right)^2 \tag{12}$$

where $\mathcal{D}$ means decoder time indices where attention predicts future values. This alignment ensures that attention responds to genuine state changes: head averaging removes spurious variation, local window compression isolates the relevant context, and standardization enables meaningful temporal correlation measurement, restricted to decoder positions where predictive attention operates.

**Training Process**

OmniTFT is trained on a single NVIDIA H100 Tensor Core GPU (SXM5, 80 GB) based on the Hopper architecture (compute capability 9.0). The final configuration uses a model width of 128 units, 4 stacked encoder-decoder blocks, 6 attention heads, and a dropout rate of 0.3. Training optimizes a quantile regression objective with Adam, using a learning rate of 1×10⁻⁵, a mini-batch size of 64, gradient clipping at 1.0, and a maximum of 300 epochs. Early stopping is employed with a patience of 10. Hyperparameter optimization was performed via a grid search across seven groups. Selection was performed using a single-label proxy task over 288 training iterations, and finally, the current hyperparameter combination was obtained.

# Data availability

The datasets used in this study are publicly available through PhysioNet (https://physionet.org/) upon completion of required training and approval:

**MIMIC IV v2.2** (Medical Information Mart for Intensive Care IV): Available at https://physionet.org/content/mimiciv/2.2/. This database contains deidentified health data from Beth Israel Deaconess Medical Center (2008–2019). **MIMIC III v1.4** (Medical Information Mart for Intensive Care III): Available at https://physionet.org/content/mimiciii/1.4/. This database contains deidentified health data from Beth Israel Deaconess Medical Center (2001–2012). **eICU CRD v2.0** (eICU Collaborative Research Database): Available at https://physionet.org/content/eicucrd/2.0/. This multicenter database contains deidentified health data from 208 hospitals across the United States (2014–2015). Access to these databases requires completion of the CITI "Data or Specimens Only Research" training course and signing of a data use agreement. All datasets are governed by the Health Insurance Portability and Accountability Act (HIPAA) and comply with deidentification standards. No additional institutional review board approval was required for this study as all data are publicly available and deidentified.

## Code availability

The relevant code for OmniTFT can be found in the Git repository at the following link: https://github.com/WanzheUTokyo/OmniTFT-Fais.

# Acknowledgements

Computational resources were provided by the supercomputer system SHIROKANE at the Human Genome Center, Institute of Medical Science, University of Tokyo.

# Funding

This work was supported by JST SPRING, grant number JPMJSP2108.

# Author information

W.X. contributed to the data preprocessing, model design, and formal analysis. W.X, Y.D contributed to the research design. Y.D. contributed to the original data, analysis design. W.X. contributed to the original draft. M.L., W.Z, Y.C. Y.Y., X.Z., S.J.P and K.N. contributed to the reviewing and editing. K.N. supervised the study. All the authors have read and approved the final version of the manuscript.

# Declarations

**Ethics approval and consent to participate**

Not applicable.

**Consent for publication**

Not applicable.

**Competing of interests**

The authors declare that they have no competing interests.

# Tables:

**Table A-1. Summary of MAE (MAPE%) across internal average and external validation sets**

| Label | MIMIC IV test (%) | MIMIC III (%) | eICU (%) |
| --- | --- | --- | --- |
| Blood Pressure | 5.05 (6.67) | 6.47 (8.60) | 3.98 (4.34) |
| Heart Rate | 4.12 (4.68) | 6.29 (7.10) | 6.65 (7.02) |
| Creatinine | 0.30 (15.86) | 0.62 (12.92) | 0.31 (12.29) |
| SpO2 | 1.04 (1.09) | 1.90 (2.03) | 2.15 (2.33) |
| SF ratio | 6.56 (3.69) | 11.70 (6.16) | 13.21 (7.62) |
| Lactate | 0.35 (18.20) | 0.76 (26.03) | 0.61 (21.45) |
| Respiratory Rate | 1.72 (8.61) | 2.75 (13.82) | 2.28 (11.52) |
| Temperature | 0.62 (1.67) | 0.50 (1.37) | 0.75 (2.08) |

Table A-1. MAE, mean absolute error; MAPE, mean absolute percentage error (%)

**Table A-2. Comparison of OmniTFT with Other Methods**

| Method | BP | HR | SpO$_2$ | RR | Temp | Creatinine | Lactate | SF ratio |
| --- | --- | --- | --- | --- | --- | --- | --- | --- |
| Vanilla TFT | 11.78 (15.58) | 13.53 (15.37) | 21.26 (22.29) | 14.08 (70.49) | 5.24 (14.11) | 3.70 (>1) | 0.95 (49.41) | 24.50 (18.2) |
| LSTM | 18.10(23.91) | 26.28(>1) | 9.36(9.83) | 14.29(71.53) | 3.99(10.76) | 6.01(>1) | 0.67 (34.84) | 54.16(40.2) |
| Prophet | 13.80 (16.07) | 11.23 (11.54) | 1.75 (3.85) | 5.77 (16.69) | 3.71 (4.02) | 0.53 (27.76) | 0.61 (31.85) | — |
| GRU | 20.03(26.46) | 15.51(17.62) | 18.30(19.18) | 10.66(53.36) | 5.45(14.70) | 17.84(>1) | 1.96 (>1) | 53.51(39.8) |
| Nephrocast | — | — | — | — | — | **0.13 (4.55)** | — | — |
| TSMixer | 10.72 (12.62) | 11.39 (14.12) | 2.85 (5.99) | **1.18 (7.36)** | 0.82 (0.85) | 0.50 (26.18) | 0.58 (30.05) | — |
| OmniTFT (Ours) | **5.05 (6.67)** | **4.12 (4.68)** | **1.04 (1.09)** | 1.72 (8.61) | **0.62 (1.67)** | 0.30 (15.86) | **0.35 (18.20)** | **6.56 (3.69)** |

Table A-2. Each cell reports MAE (MAPE%). "—" indicates not evaluated or not reported. Formula methods are only applicable to SF ratio. Percent errors marked ">1" reflect undefined or non-informative percentage scaling for that target. OmniTFT achieves the lowest MAE on all eight targets, spanning both vital signs and laboratory results.

# Figures:

**Figure 1**

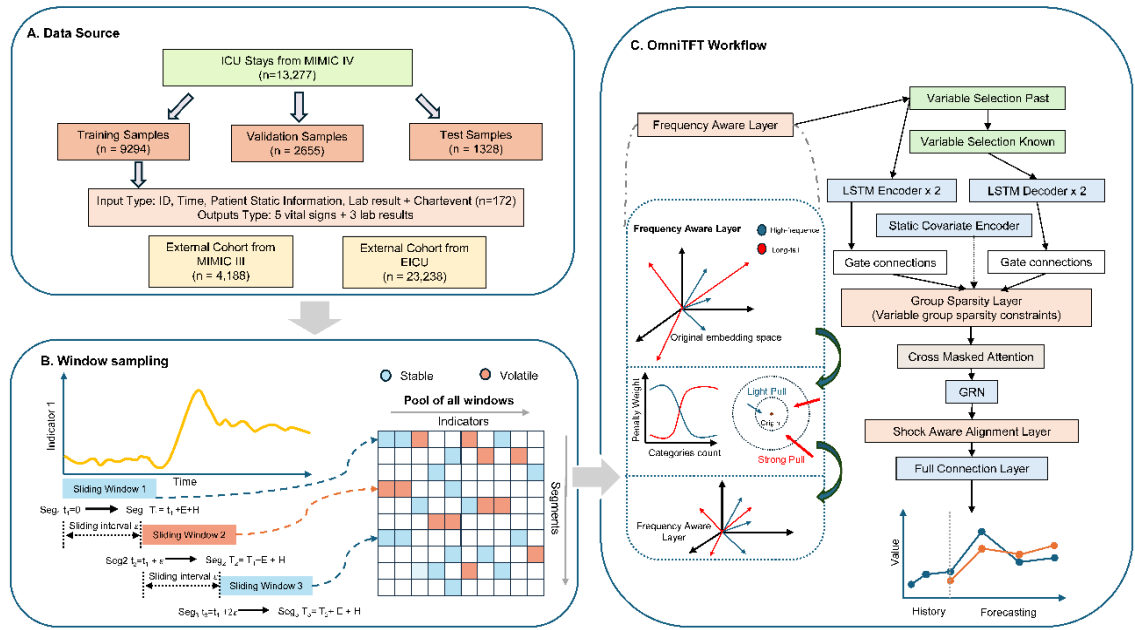

Figure 1. Workflow of OmniTFT. (A) Data Source. MIMIC-IV ICU stays (n=13,277) are split into training (n=9,294), validation (n=2,655), and test (n=1,328) sets. (B) Window Sampling. The yellow curve represents physiological indicator values over time. A two-state HMM classifies time periods into stable (blue) and volatile (red) states. Sliding windows, with encoder window E and forecast horizon H, are sampled using a balanced scheme to ensure equal representation of both states. (C) OmniTFT Workflow. Input tokens are processed through separate pathways for past variables and known future variables, with a Frequency Aware Layer stabilizing categorical embedding. Historical data flows through two stacked LSTM encoder layers, while static covariates are processed by a Static Covariate Encoder. Two LSTM decoders generate future predictions, connected via gated mechanisms. Group Sparsity Layer enforces semantic variable grouping through sparsity constraints on variable groups. Cross Masked Attention connects encoder and decoder representations. A Gated Residual Network (GRN) processes attention outputs. Shock Aware Alignment Layer calibrates attention to align with true physiological state transitions. A Full Connection Layer produces the final predictions, transitioning from observed history to forecasted values (bottom graph). The inset diagram illustrates the frequency-aware embedding mechanism: high-frequency categories (black arrows) experience light regularization pulls toward the origin, while long-tail rare categories (red arrows) undergo strong pull, preventing overfitting while maintaining embedding stability.

**Figure 2**

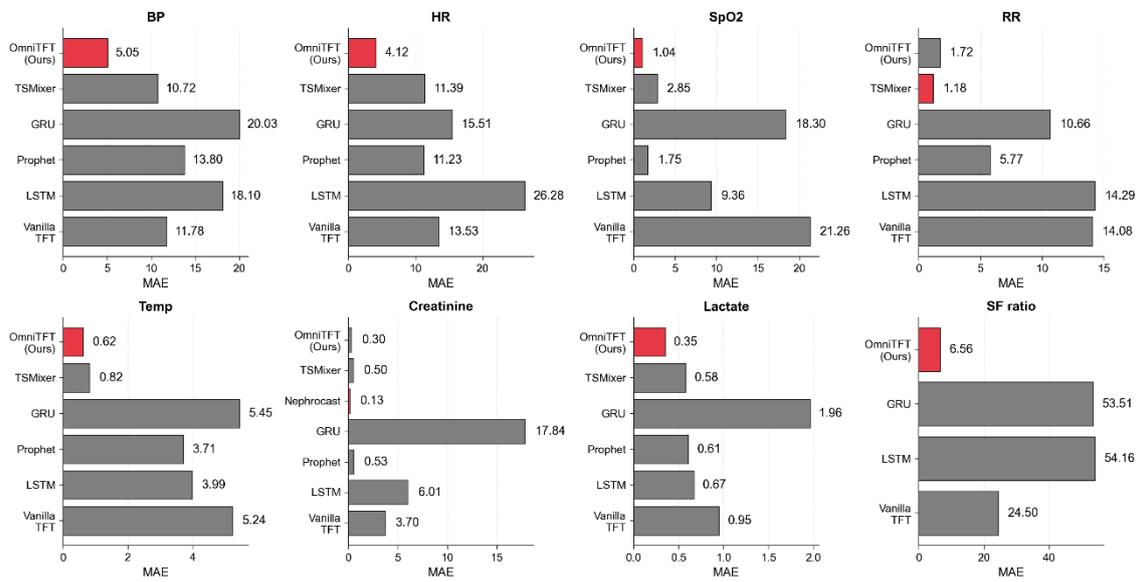

Figure 2. Performance comparison of OmniTFT and baseline models on eight clinical metrics. Each panel shows the mean absolute error (MAE) for a specific metric: blood pressure (BP), heart rate (HR), oxygen saturation (SpO2), respiratory rate (RR), temperature (Temp), creatinine, lactate, and P/F ratio. The best model is highlighted in red, while the other models are shown in gray. Lower MAE values indicate better performance.

**Figure 3**

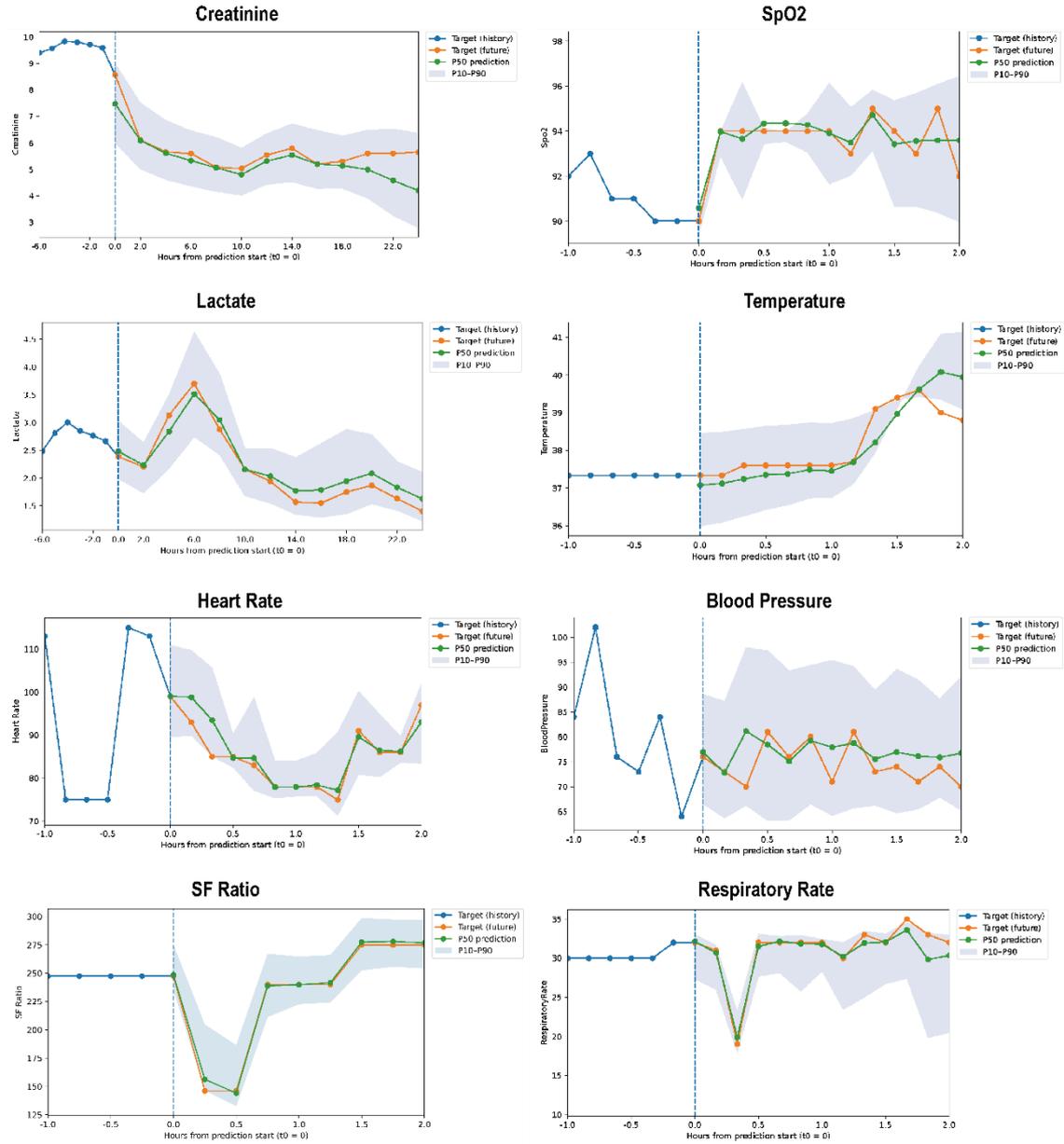

Figure 3. Representative forecasting trajectories for eight clinical indicators. Each panel shows predictions for a patient whose mean absolute error (MAE) for that indicator is closest to the cohort mean, representing a typical forecasting case. Blue lines show historical observations (Target history), orange lines show actual future values (Target future), green lines show median predictions (P50), and shaded regions show the P10-P90 prediction intervals. The vertical dashed line at t = 0 marks the prediction start point, with historical data on the left and forecasts on the right. For consistent visualization, historical windows are displayed as 6 hours for creatinine and lactate, and 1 hour for all other indicators.

Figure 4

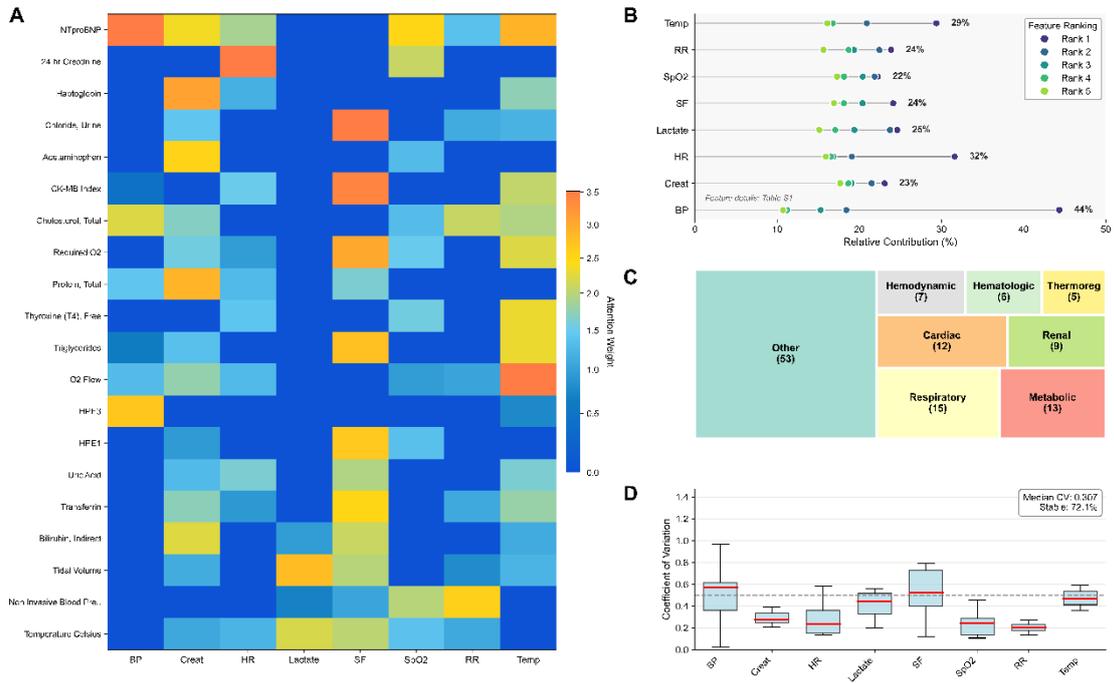

Figure 4. Analysis of multi-target attention weights for ICU vital sign prediction. (A) Heatmap of the composite attention weights of the top 20 features across eight prediction targets. Attention weights are aggregated across all encoder layers, attention heads, and cross-validation folds. BP: blood pressure; Creat: creatinine; HR: heart rate; SF: oxygenation index; RR: respiratory rate; Temp: temperature. Color intensity indicates attention strength. (B) Dot plot of the relative contributions of the top five features for each prediction target. Each dot represents a feature, positioned by contribution percentage and colored by rank (Rank 1-5). The names and contribution percentages of the top five features are provided in Supplementary Information Table S4-1. Lines connect points within the same target to visualize the contribution gradient. The contribution of the top feature is indicated by percentage. Gray reference lines extend from zero to the maximum contribution value for each target. (C) Physiological system distribution of highly attentive features. The area of the rectangle represents the feature count for that category (shown in parentheses). The top 15 features across all eight prediction targets were assigned categories based on clinical domain keywords. (D) Boxplot of attention stability for the top 10 features for each predicted target. Measured by the coefficient of variation (CV = standard deviation/mean). Boxplots show the median (red horizontal line), interquartile range (box boundaries), and the data range excluding outliers (whiskers). The horizontal dashed line marks the stability reference threshold of CV = 0.5. The x-axis represents the predicted target, and the y-axis represents the CV.

**Figure 5**

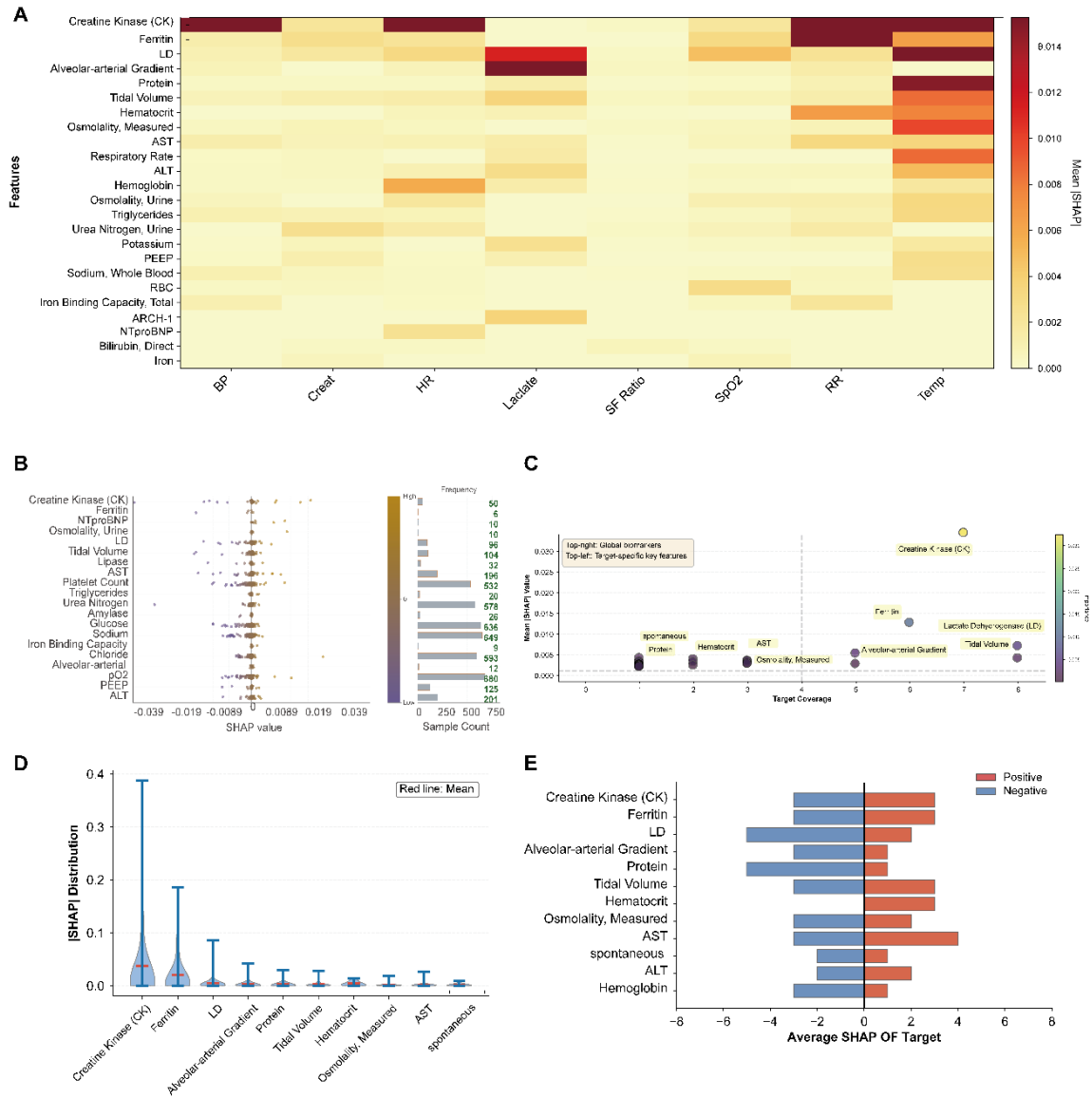

Figure 5. SHAP analysis reveals feature importance patterns and directional contributions across prediction targets. (A) Feature-target heatmap showing mean SHAP values for each feature across eight prediction targets (BP: blood pressure, Creat: creatinine, HR: heart rate, Lactate, SF Ratio: SpO2/FiO2 ratio, SpO2: oxygen saturation, RR: respiratory rate, Temp: temperature). Color intensity indicates contribution magnitude. (B) Beeswarm plot showing patient-level SHAP value distributions for heart rate prediction. Each dot represents one patient, with horizontal position indicating SHAP value, dot color encoding feature value magnitude, and the right histogram showing sample frequency. (C) Feature classification by target coverage (x-axis) versus mean SHAP value (y-axis). Top-right quadrant contains global biomarkers affecting ≥6 targets, left contains target-specific features with high importance but narrow coverage. (D) Violin plots displaying |SHAP| value distributions across all samples for top features. Width indicates distribution density; red line marks the mean. (E) Directional decomposition of average SHAP values across targets. Blue bars represent negative contributions (decreasing predictions), red bars represent positive contributions (increasing predictions). Symmetry indicates context-dependent bidirectional effects.